\documentclass[letterpaper]{article} 
\usepackage{aaai25}  
\usepackage{times}  
\usepackage{helvet}  
\usepackage{courier}  
\usepackage[hyphens]{url}  
\usepackage{graphicx} 
\usepackage{natbib}  
\usepackage{caption} 
\usepackage{algorithm}
\usepackage{algorithmic}

\usepackage{newfloat}
\usepackage{listings}
\DeclareCaptionStyle{ruled}{labelfont=normalfont,labelsep=colon,strut=off} 
\floatstyle{ruled}
\newfloat{listing}{tb}{lst}{}
\floatname{listing}{Listing}
\usepackage{graphicx}
\usepackage{xcolor}
\usepackage{subfiles}
\usepackage{multirow}
\usepackage{subcaption}
\usepackage{dashrule}
\usepackage{enumitem} 
\usepackage{booktabs} 
\usepackage{colortbl} 
\usepackage{pifont} 
\usepackage{subfiles}
\usepackage{amssymb}
\usepackage{amsmath} 
\usepackage{multicol}
\title{MAMS: Model-Agnostic Module Selection Framework for Video Captioning}
\author{
    Sangho Lee\textsuperscript{\rm 1,\rm2},
    Il Yong Chun\textsuperscript{\rm 1,\rm 3}\thanks{Corresponding authors.},
    Hogun Park\textsuperscript{\rm 1}\footnotemark[1] \\
}
\affiliations {
    \textsuperscript{\rm 1}Sungkyunkwan University, Suwon, Republic of Korea\\
    \textsuperscript{\rm 2}Hippo T\&C Company Limited, Suwon, Republic of Korea\\
    \textsuperscript{\rm 3}Center for Neuroscience Imaging Research, Institute for Basic Science, Suwon, Republic of Korea\\
    lsh3982@hippotnc.com, \{iychun, hogunpark\}@skku.edu
}

\usepackage{bibentry}

\begin{document}

\maketitle

\begin{abstract}
Multi-modal transformers are rapidly gaining attention in video captioning tasks.
Existing multi-modal video captioning methods typically extract a fixed number of frames, which raises critical challenges.
When a limited number of frames are extracted, important frames with essential information for caption generation may be missed.
Conversely, extracting an excessive number of frames includes consecutive frames, potentially causing redundancy in visual tokens extracted from consecutive video frames.
To extract an appropriate number of frames for each video, this paper proposes the first model-agnostic module selection framework in video captioning that has two main functions: 
\textit{(1)} selecting a caption generation module with an appropriate size based on visual tokens extracted from video frames, and 
\textit{(2)} constructing subsets of visual tokens for the selected caption generation module.
Furthermore, we propose a new adaptive attention masking scheme that enhances attention on important visual tokens.
Our experiments on three different benchmark datasets demonstrate that the proposed framework significantly improves the performance of three recent video captioning models.
\end{abstract}

%

\section{Introduction}\label{sec:intro}

The video captioning task generates descriptions for provided videos in natural language~\citep{li2021value,wang2019controllable}.
It has been rapidly gaining attention in blind navigation, video event commentary, human-computer interaction, etc.
To improve video captioning performances, it is pivotal to introduce multi-modal transformers~\citep{sun2019videobert}.
Many recent studies extract an identical number of frames for different videos to use a consistent input size for transformer-based models; see, e.g., \citep{chen2023vast,yang2023vid2seq}. 

\begin{figure}[!t]
  \centering
  \includegraphics[width=0.45\textwidth]{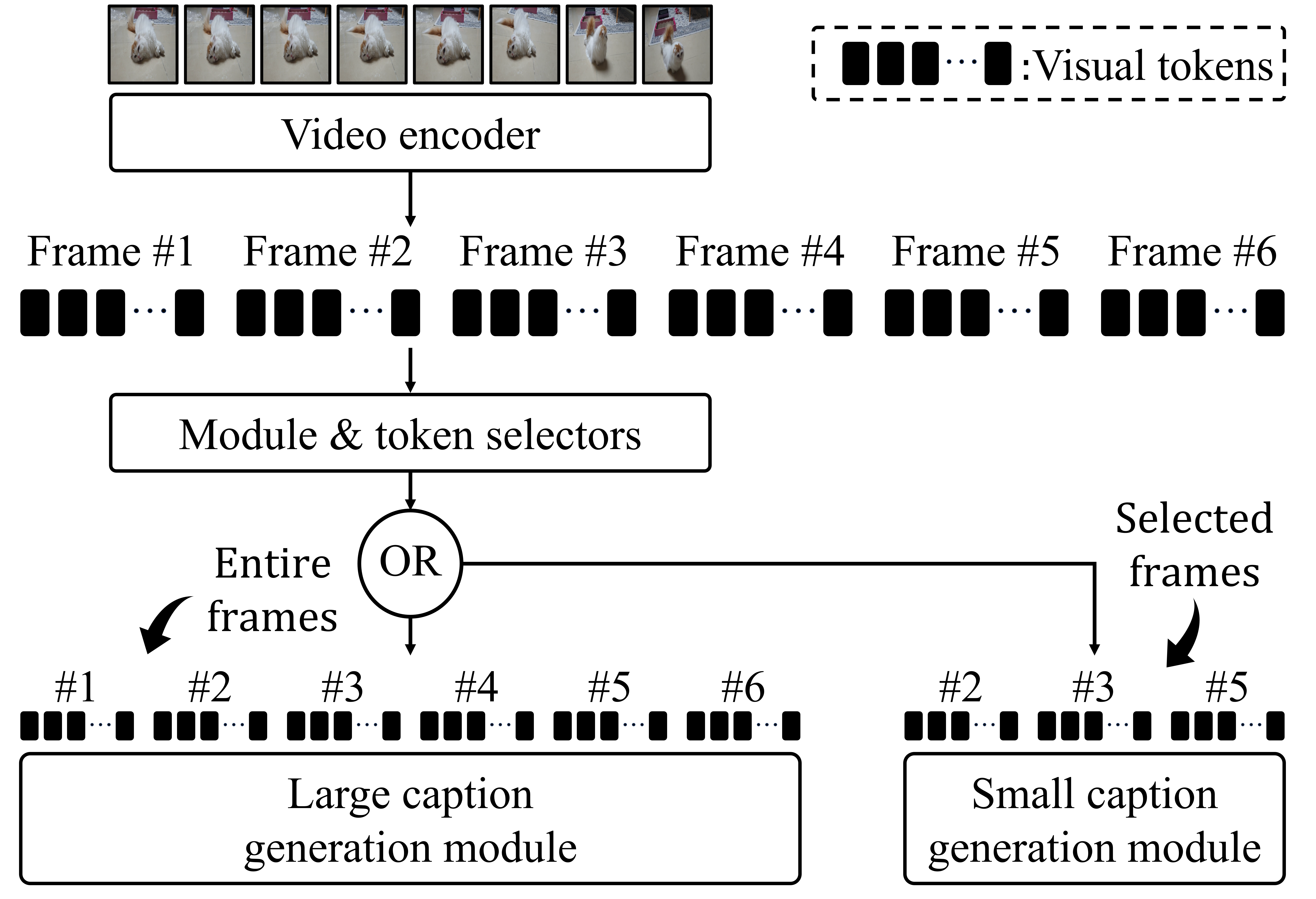}
  \caption{The overview of the proposed framework.}
  \label{fig:introfig}
\end{figure}
Selecting a fixed number of frames in existing captioning models has critical limitations.
For videos with abundant information, e.g., videos with large dynamics, if we extract a limited number of frames, caption generation performances can degrade~\citep{lin2022swinbert}.
It may omit frames that encapsulate essential information for caption generation, potentially compromising the accuracy and completeness of caption generation~\citep{Gao_2022_ACCV}.
Conversely, for videos with little information, e.g., videos with little dynamics, if frames are densely extracted, it could result in extracting similar frames.
Numerous studies have demonstrated that redundant visual tokens generated by a large number of consecutive frames can adversely affect performance~\citep{Ada-Vit,liang2022evit}.
Our observations indicate that the performance of existing video captioning methods stagnates or even declines as the number of frames increases.
We conjecture that this phenomenon is caused by the fact that existing methods rely on a fixed number of frames across all videos.
This study assumes in caption generation that it could be more reasonable to vary the number of frames or visual tokens for each video, rather than to fix it for all videos.

To address the aforementioned issues, this paper proposes the first \textbf{Model-Agnostic Module Selection (MAMS)} framework in video captioning that adaptively selects a caption generation module for each video, where each module extracts a different number of frames.
The proposed framework consists of an existing caption generator that uses all frames, a smaller caption generator module that uses a subset of frames, and a module \& token selector that selects the appropriate generation module and tokens for each video, respectively.
Figure~\ref{fig:introfig} illustrates an overview of the proposed framework.
The process of the proposed framework is given as follows. 
\textit{(1)} We extract visual tokens from video with a video encoder.
\textit{(2)} Using the proposed module \& token selectors, we select an appropriate size of a generation module. 
If a smaller module is selected, we then construct a subset of visual tokens corresponding to selected frames from a full set of visual tokens.
\textit{(3)} We input selected visual tokens -- combined with textual tokens -- to either a large or small caption generation module.
Different from existing models that use a fixed number of frames and visual tokens, the proposed framework adaptively selects a caption generation module with an appropriate size, which results in using a varying number of frames and visual tokens for each video.
Moreover, we introduce a new adaptive attention masking scheme that focuses more on visual tokens with higher contributions to caption generation.
Ultimately, we select/focus on essential visual tokens in both frame and token levels.

The contributions of the paper are summarized as follows:
\begin{itemize}[leftmargin=*]
\item \textbf{(Problem discovery)} We discover a performance saturation problem in existing captioning methods that extract the same number of frames from all videos.
\item \textbf{(New methodology)} We propose the first \textbf{MAMS} framework in video captioning that selects an appropriate caption generation module and important visual tokens for each video in terms of frame level.
Additionally, we propose a new \textbf{adaptive attention mask} that focuses more on important visual tokens.
\item \textbf{(Broad applicability \& performance improvement)} Our framework is applicable to existing video captioning models and effectively addresses their limitations, significantly improving the captioning performance of the three state-of-the-art models, SwinBERT, UniVL, and mPLUG-2. Notably, applying the MAMS framework to mPLUG-2 achieved a new state-of-the-art benchmark.
\end{itemize}

\section{Related Works}
\subsection{Video Captioning}
Early video captioning approaches used rule-based methods, directly extracting subjects, verbs, and objects to construct sentences~\citep{das2013thousand,kojima2002natural}.
Subsequent studies extracted sentences on a frame-by-frame basis and combined them~\citep{bahdanau2014neural,sutskever2014sequence}.
The majority of recent models are based on multi-modal transformers, which simultaneously utilize visual tokens extracted from videos and textual tokens from pre-generated words to generate sentences~\citep{arnab2021vivit, sun2019videobert}.
Initially, these approaches generated sentences using pre-extracted visual tokens~\citep{aafaq2019spatio,pan2020spatio,pei2019memory,shi2020learning}.
Over time, approaches using pre-extracted tokens have advanced into an end-to-end framework that directly extracts visual tokens from raw videos to generate sentences.
This end-to-end approach significantly enhances performance, as the visual token extractor can be jointly optimized with the caption generation module~\citep{lin2022swinbert}.
{We categorize the multi-modal transformer based end-to-end} approach into two main classes.
The first category is called sparse sampling that selects a limited number of frames from the video~\citep{fu2023empirical,wang2022language}. 
This approach could miss important information needed in generating captions for some videos with large dynamics.
The second category is called dense sampling that extracts visual tokens with a sufficient number from the consecutive video frames~\citep{kuo2023mammut,xu2023mplug,lin2022swinbert}. 
This approach could generate redundant visual tokens for some videos with small dynamics that negatively affect caption generation performance.
This paper proposes a new framework that can overcome the limitations of existing caption generation models using a fixed number of frames.

\subsection{Limitation of a Fixed Learnable Attention Mask}

SwinBERT~\citep{lin2022swinbert} focuses more on important tokens for caption generation by introducing a learnable attention mask that is consistently applied to all videos.
This approach demonstrates improved performance by paying more attention to visual tokens corresponding to the center of each frame compared to those corresponding to the edges of each frame.
However, a fixed learnable attention mask has two limitations as follows.
\textit{(1)} It reduces attention values at the edges of frames across all videos.
Suppose important parts for caption generation are located at the edges of the frame in a video. 
In that case, these parts can be missed in caption generation, leading to performance degradation~\citep{lin2022swinbert}.
\textit{(2)} Additionally, a fixed learnable attention mask cannot consider the unique characteristics of each video. 
The proposed adaptive attention masking scheme can overcome the limitations in the existing fixed learnable attention mask scheme.

\section{	Preliminary Analysis}\label{Sec:Preliminary}

\begin{figure}[t!]
  \centering
  \includegraphics[width=0.4\textwidth]{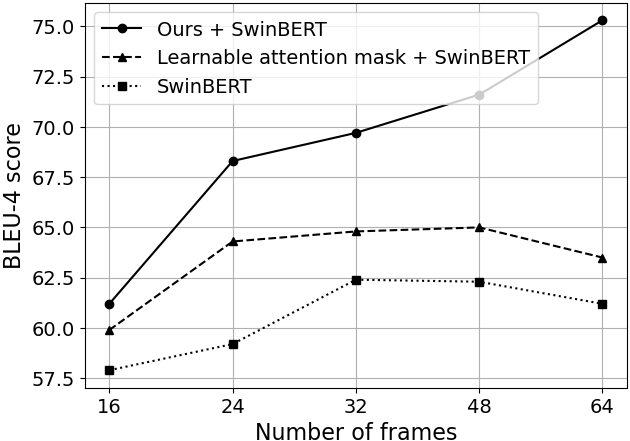}
  \caption{BLEU-4 scores~\citep{papineni2002bleu} with different numbers of video frames in SwinBERT (the MSVD datasets).
  The dotted, dashed, and solid lines represent different experiments involving SwinBERT, with and without an attention mask, and the proposed MAMS framework.}
    \label{fig:motivation_flowchart}
\end{figure}

This section presents the experimental analysis of the aforementioned saturation issue.
The dotted and dashed lines in Figure~\ref{fig:motivation_flowchart} show the performance of the SwinBERT~\citep{lin2022swinbert} model without and with a fixed learnable attention mask, respectively.
We observed with the existing methods that the captioning performance improves as the number of frames increases, but gets saturated or even degraded beyond a certain number of frames.
Increasing the number of frames, in general, can use crucial frames and improve caption generation performance.
However, extracting too many frames can lead to redundancy in visual tokens, resulting in performance degradation.
We argue that for accurate caption generations, one needs to vary the number of frames to be extracted for different videos.
These experimental results motivate the proposed MAMS framework that can vary the number of frames and visual tokens for each video captioning.
The solid lines in Figure~\ref{fig:motivation_flowchart} show that, by using the proposed \textbf{MAMS} framework, we achieve consistent performance improvement as the number of frames increases, effectively addressing the limitations of using a fixed number of frames, i.e., visual tokens, in existing models.

\section{Methods}

\begin{figure}[t!]
    \centering
    \includegraphics[width=0.40\textwidth]{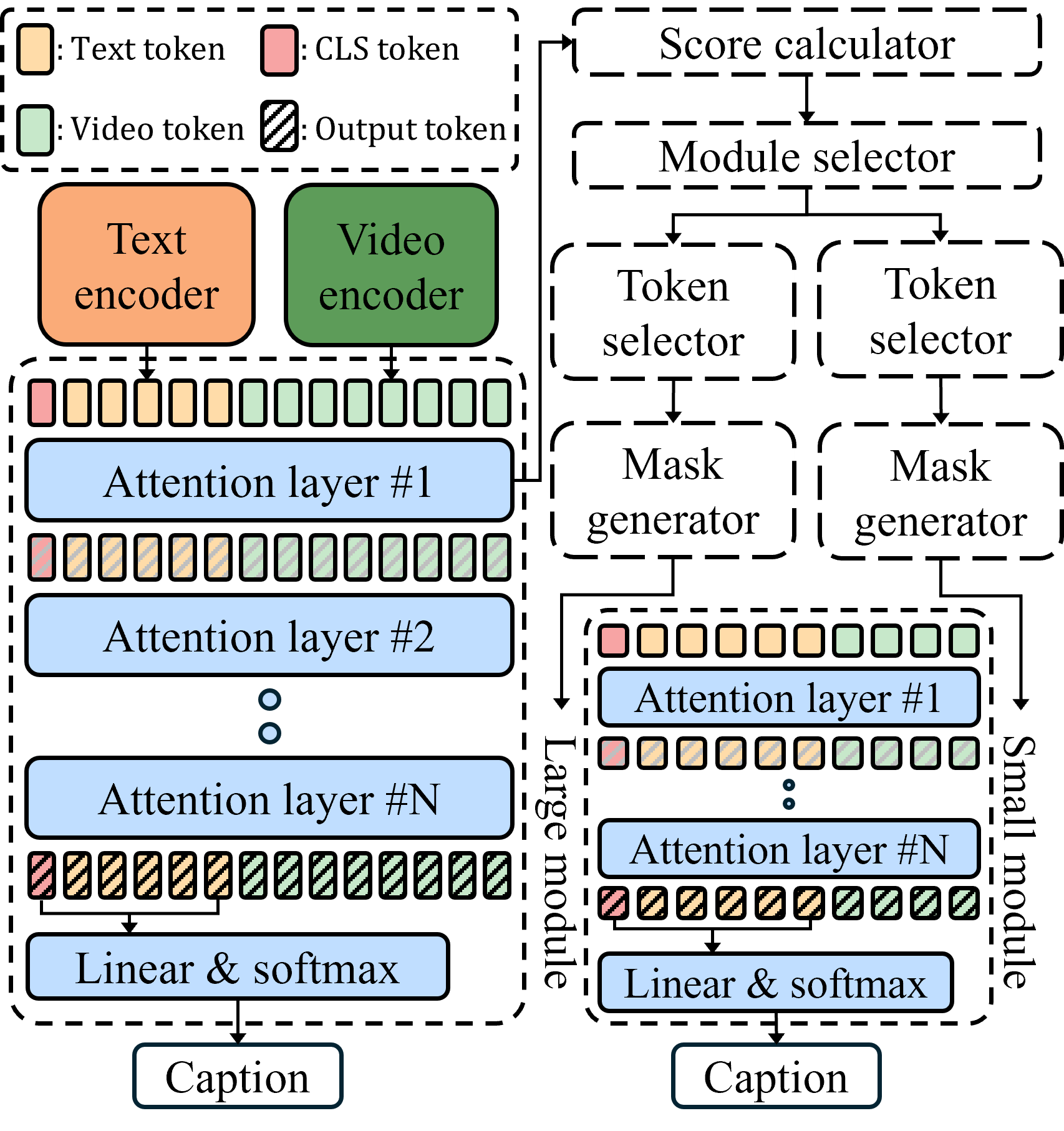}
    \caption{The overall MAMS framework}
    \label{fig:overall}
\end{figure}

\subsection{The Overall Architecture of MAMS Framework}

In video captioning, many popular architectures based on multi-modal transformers consist of three major modules:
\textit{1)} a video encoder that transforms a video into visual tokens;
\textit{2)} a text encoder that transforms a caption into textual tokens; and
\textit{3)} a caption generation module that creates captions.
In a nutshell, the proposed \textbf{MAMS} framework augments the aforementioned architecture by introducing a smaller caption generation module in parallel.
We differentiate two generation modules with the terms, a \emph{large} and a \emph{small} module, which are tailored for inputs of sizes $T_\text{large}$ and $T_\text{small}$, respectively.
Furthermore, our MAMS framework includes a \emph{score calculator} for calculating the importance of each visual token, referred to as a token significance score.
Based on this score, \emph{module and token selectors} within the MAMS framework strategically choose between two modules for optimal training and inference. 
Additionally, a \emph{mask generator} of the MAMS framework creates an \textbf{adaptive attention mask} for each attention layer, based on token significance scores.
Figure~\ref{fig:overall} shows the overall architecture of the MAMS framework.

\subsection{Token Significance Score}\label{sec:token_significance_score_section}

In video captioning models, a video encoder transforms frames into visual tokens.
A caption generation module then takes these visual tokens to produce captions. 
As adjacent frames are similar, it is natural that their visual tokens have similar values. 
We assume, however, that their contributions to caption generation are different.

To quantify the contribution of each token to caption generation,
we define a token significance score inspired by~\citep{cao2023ada}.
Specifically, we define the token significance score of the $p$th token at the $i$th frame as follows:
\begingroup
\begin{equation}
\label{eq:token-sig}
t_{i,p} = \frac{a_{i,p} \cdot \|\mathbf{x}^\text{v}_{i,p}\|}{\sum_{i,p} a_{i,p} \cdot \|\mathbf{x}^\text{v}_{i,p}\|}, \quad i = 1,\ldots,T_\text{large}, ~p = 1,\ldots,P,
\end{equation}
\endgroup
where \(a_{i,p}\) denotes the attention value between a special classification (CLS) token and a $p^\text{th}$ visual token at the $i^\text{th}$ frame, 
\(\mathbf{x}^\text{v}_{i,p}\) denotes a $p^\text{th}$ visual token at a $i^\text{th}$ frame,
$T_\text{large}$ is the number of total frames, and $P$ is the number of visual tokens per video frame.
We calculate $\{ t_{i,p} : \forall i,p \}$ from the first attention layer of a caption generation module.
Considering a CLS token as representing the starting point of a caption, 
Attention values between a CLS token and visual tokens can quantify the contribution of visual tokens to the entire caption~\citep{cao2023ada}.
In Eq. \eqref{eq:token-sig}, we additionally assume that not only the attention values but also the visual tokens themselves influence caption generation and use the norm values of each visual token in computing the token significance scores. 

\subsection{Module and Token Selector}

\begin{figure}
  \centering
  \includegraphics[width=0.33\textwidth]{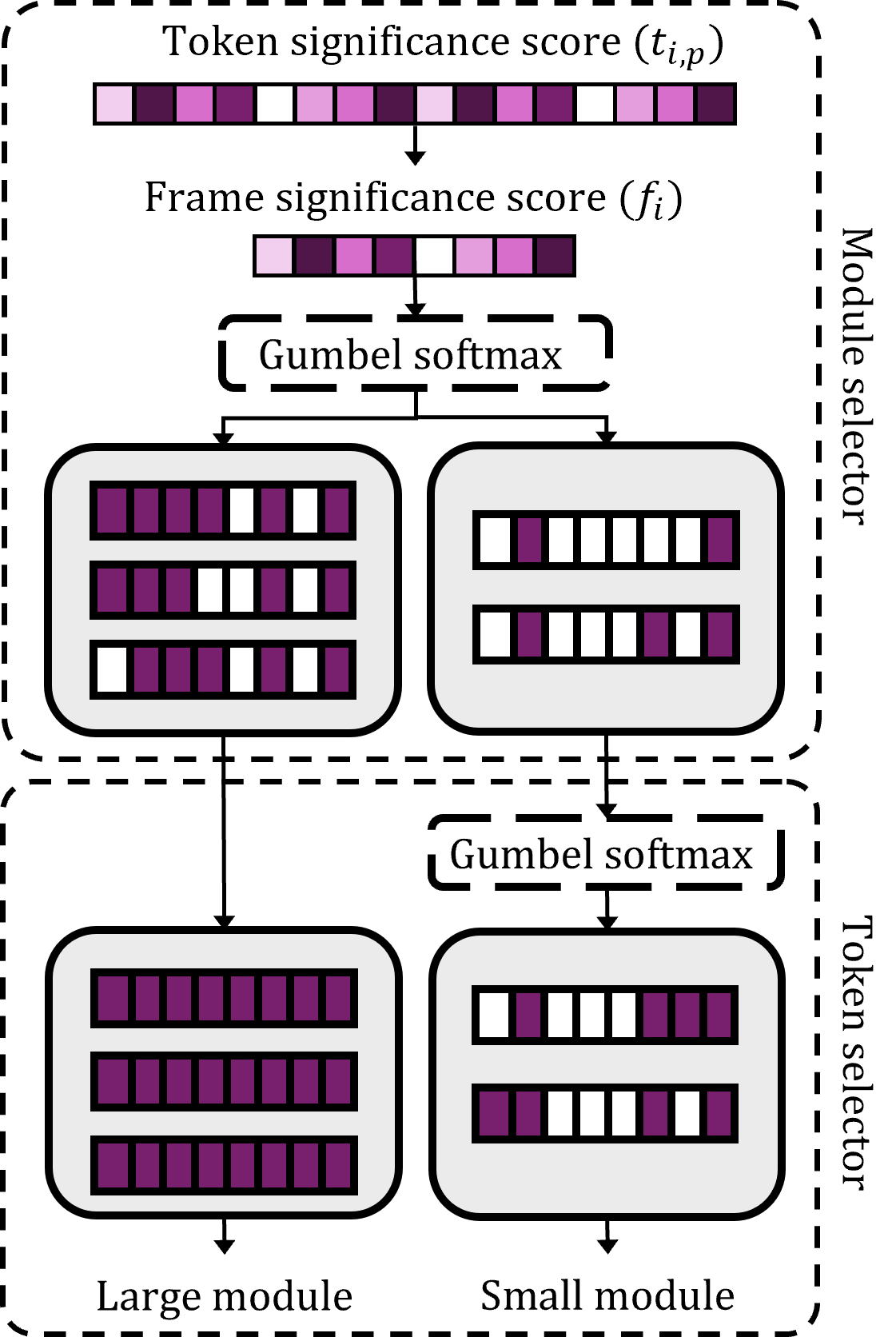}
  \caption{Illustration of proposed module and token selector}
    \label{fig:module and token selector}
\end{figure}

\subsubsection{Module Selector}
Using calculated $\{ t_{i,p} : \forall i,p \}$ in Eq. (\ref{eq:token-sig}), we define a frame significance score for the $i$th frame as follows:
\begingroup
\begin{equation}
\label{eq:frm-sig}
f_i = \sum_{p=1}^P t_{i,p}, 
\quad i = 1,\ldots,T_\text{large},
\end{equation}
\endgroup
where by default, we consider that a video encoder generates multiple visual tokens from a single frame.
Calculating the defined quantities Eqs. (\ref{eq:token-sig})--(\ref{eq:frm-sig}), we use them to select important frames for caption generation module selection.

We first select important frames,
applying a \texttt{for} loop algorithm based on the Gumbel-Softmax operator~\citep{jang2016categorical} to $\{ f_i : i = 1, \ldots, T_{\text{large}} \}$.
As we run the Gumbel-Softmax operator $T_\text{large}$ times,
the same frame may be selected multiple times so the number of selected frame indices can vary for different videos.
In choosing between small and large caption generation modules, we apply the following selection rule using the set of selected frame indices $S^\text{frm}$:
\begingroup
\begin{equation}
\label{eq:mod-select}
\left\{ 
\begin{array}{ll} 
\text{Select~a~small~module}, & \text{if}~ |S^\text{frm}| \leq T_\text{small},
\\
\text{Select~a~large~module}, & \text{if}~ |S^\text{frm}| > T_\text{small},
\end{array} \right.
\end{equation}
\endgroup
where $|S^\text{frm}|$ denotes the number of selected frame indices.
The decision rule in Eq. (\ref{eq:mod-select}) implies the following.
The condition $|S^\text{frm}| > T_{\text{small}}$ implies that a small module may miss important video frames for caption generation, needing to use a large module.
The condition $|S^\text{frm}| \leq T_\text{small}$ suggests that the selected frames adequately contain the important frames for caption generation, leading to selecting a small module.
The \emph{Module selector} in Figure ~\ref{fig:module and token selector} illustrates the above module selection process.

\subsubsection{Token Selector}

If a small module is selected as an appropriate one, i.e., $|S^\text{frm}|$ is less than or equal to $T_\text{small}$,
we construct a final set of frame indices as follows.
We apply a \texttt{while} loop algorithm based Gumbel-Softmax operator until the number of selected indices reaches $T_\text{small}$ for a small module.
We use \texttt{while} loops algorithm to ensure that the input sizes match the requirements of a small generation module.
(See details of the Gumbel-Softmax-based algorithm in the supplementary material.)
Conversely, if $|S^\text{frm}|$ exceeds $T_\text{small}$, we use all frame indices for a large module.
Through this process, we create distinct sets of visual tokens for each video. 
Finally, these selected visual tokens are concatenated with textual tokens to construct the input for each module.
The \emph{Token selector} in Figure ~\ref{fig:module and token selector} illustrates the above token selection process.

\subsection{Adaptive Attention Mask}

\begin{figure}
  \centering
  \includegraphics[width=0.43\textwidth]{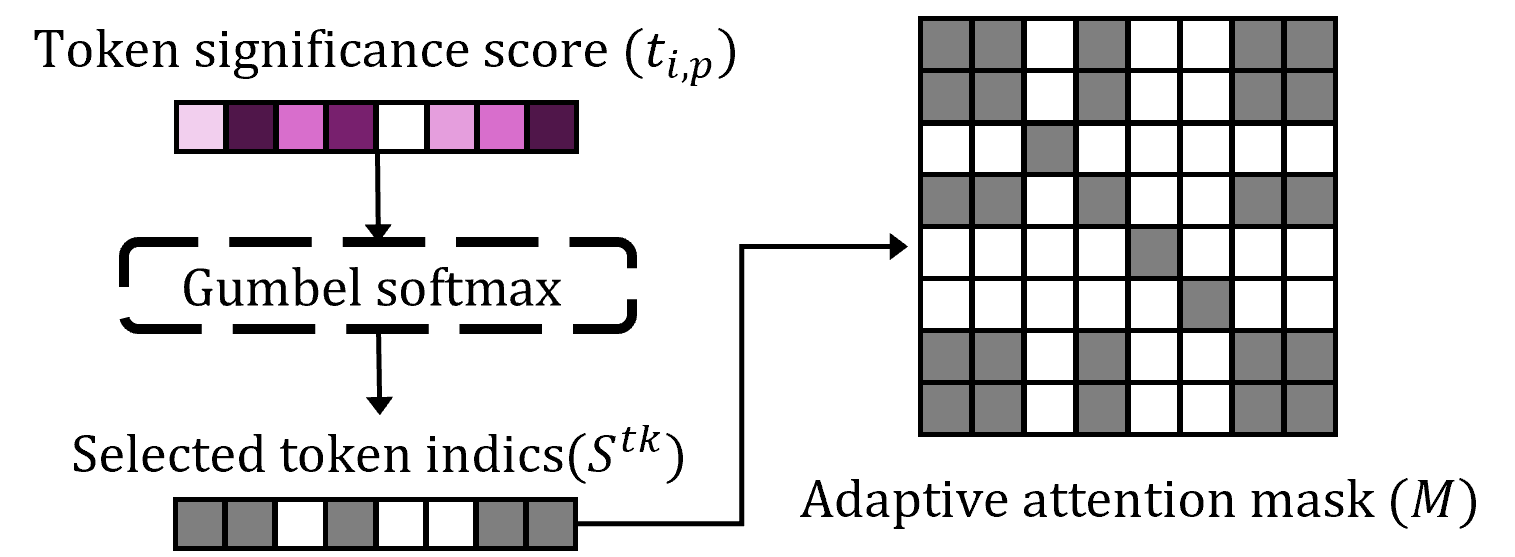}
  \caption{Illustration of the proposed adaptive attention masking scheme}
    \label{fig:AdaptiveMask}
\end{figure}

A token selector in the proposed MAMS framework selects appropriate visual tokens at the frame level based on their contribution to caption generation.
Even though visual tokens are generated from the identical frame, their contributions to caption generation may vary.
To better select important visual tokens from selected or all frames, we propose a new adaptive attention masking scheme, where the corresponding binary mask for each frame of a small and large module is denoted as \(\mathbf{M}_\text{small}\) and $\mathbf{M}_\text{large}$, respectively.
We use the adaptive attention mask to enable each module in the proposed MAMS framework to focus more on visual tokens with a higher contribution in caption generation.
Figure~\ref{fig:AdaptiveMask} illustrates the proposed adaptive attention masking scheme.

First, we extract the indices of important tokens based on their contribution to caption generation and use those indices to generate \(\mathbf{M}_\text{large}\).
We apply the \texttt{for} loop algorithm based on the Gumbel-Softmax operator to $\{ t_{i,p} : \forall i,p \}$ in (\ref{eq:token-sig}) to generate a set of indices of selected tokens in the form of $(i,p)$, denoted as $S^\text{tk}$.
We run the Gumbel-Softmax operator for \(T_\text{large} \cdot P\) times that represents the total number of visual tokens.
As a result, some token(s) may be selected multiple times, leading to a variable number of selected token indices across different videos, with an adaptive attention mask varying from video to video.
The elements of \(\mathbf{M_\text{large}}\) are determined using \(S^\text{tk}\) through the following process:
\begingroup
\begin{equation}
\label{eq:attention-mask}
M_\text{large}^{(\mathbf{x},\mathbf{y})} = 
\left\{
\begin{array}{ll}
1, & \text{if~} \mathbf{x} = \mathbf{y}, \\
1, & \text{if~} \mathbf{x} \neq \mathbf{y}, \mathbf{x} \in S^\text{tk}, \mathbf{y} \in S^\text{tk}, \\
0, & \text{otherwise},
\end{array}
\right.
\end{equation}
\endgroup
\noindent 
where $\mathbf{x}, \mathbf{y} \in \{(i,p) : i = 1,\ldots,T_\text{large}, p = 1,\ldots,P \}$.
We apply this masking scheme to all attention layers of a large caption generation module (if selected).
If a small caption generation module is selected, we construct $\mathbf{M}_\text{small}$ by using a subset of $\{ M_\text{large}^{(\mathbf{x},\mathbf{y})} : \forall \mathbf{x}, \mathbf{y} \in S^\text{tk} \}$ in Eq.~(\ref{eq:attention-mask}), with the indices of the visual tokens selected in the earlier token selector.
We explain further details in the supplementary material.

While a module and token selector in the MAMS framework selects an appropriate number of visual tokens at the \emph{frame} level,
the adaptive attention masking scheme focuses on more important visual tokens at the \emph{token} level.
Combined with the adaptive attention masking scheme, 
the proposed MAMS framework selects/focuses on essential visual tokens at both the frame and token levels.

\subsection{Training Phase}
For the training phase, we train both a large and small module using two loss functions in the following form:

\begin{equation}
\label{eq:training-process}
\begin{aligned}
\mathcal{L} &= \lambda_{\text{large}}\mathcal{L}_{\text{large}}
+ \lambda_{\text{small}}\mathcal{L}_{\text{small}}, \\[6pt]
(\lambda_{\text{large}}, \lambda_{\text{small}}) &=
\begin{cases}
(1,0), & \text{if } |S| > T_{\text{small}}, \\[6pt]
(0,1), & \text{if } |S| \leq T_{\text{small}}
\end{cases}
\end{aligned}
\end{equation}

 

\noindent where $\mathcal{L}_\text{large}$ and $\mathcal{L}_\text{small}$ are losses for training a large and small caption generation module, respectively.
By setting the module selection weighting parameters $(\lambda_\text{large}, \lambda_\text{small})$ as in Eq. (\ref{eq:training-process}), we nullify either small or large module training, ensuring that meaningful back propagation flows through only one module.

\section{Experimental Results and Discussion}

\subsection{Experimental Setups}
We ran experiments with three different datasets: MSVD ~\citep{msvd}, MSRVTT~\citep{msrvtt}, and YOUCOOKII datasets~\citep{youcook}.
We incorporated the following video captioning models into the proposed MAMS framework:
\begin{itemize}[leftmargin=*]
\item Two representative models: SwinBERT~\citep{lin2022swinbert} and UniVL~\citep{luo2020univl}
\item The state-of-the-art model, mPLUG-2~\citep{xu2023mplug}. The mPLUG-2 model is the state-of-the-art video captioning model, particularly for the MSVD and MSRVTT datasets.
\item While the SwinBERT and M-PLUG-2 models extract visual features directly from raw videos for training, the UniVL model trains on pre-extracted features.
The UniVL model, unlike the other two models, is a modular (i.e., non-end-to-end) approach.
\end{itemize}

We evaluated the generated captions using four different evaluation metrics: BLEU-4~\citep{papineni2002bleu}, METEOR~\citep{banerjee2005meteor}, ROUGE~\citep{rouge}, CIDEr~\citep{vedantam2015cider}. 
Throughout the tables, we denote the above metrics as B4, M, R, and C, respectively.
For our experiments, we used PyTorch~\citep{paszke2019pytorch} and NVIDIA A100 GPUs.
See details of experiments and implementation in the supplementary material.

\subsection{Main Results}

\begin{table}[t!]
    \centering
    
    \begin{subtable}{\columnwidth}
        \centering
        \scalebox{0.83}{
            \begin{tabular}{l cccc}
                \toprule[1.5pt]
                Models & B4 & M & R & C \\  
                \midrule
                *TextKG ~\citep{gu2023text} & 60.8 & 38.5 & 75.1 & 105.2 \\
                *CoCap ~\citep{shen2023accurate} & 60.1 & 41.4 & 78.2 & 121.5 \\
                *VIOLETv2 ~\citep{fu2023empirical} & - & - & - & 139.2 \\
                \midrule
                SwinBERT~\citep{lin2022swinbert} & 58.2 & 41.3 & 77.5 & 120.6 \\
                MAMS + SwinBERT & \textbf{60.9} & \textbf{42.1} & \textbf{78.9} & \textbf{125.0} \\
                & \textcolor{blue}{(+2.7)} & \textcolor{blue}{(+0.8)} & \textcolor{blue}{(+1.4)} & \textcolor{blue}{(+4.4)} \\
                \midrule
                mPLUG-2~\citep{xu2023mplug} & 75.0 & 48.4 & 85.3 & 165.8 \\
                MAMS + mPLUG-2 & \textbf{76.9} & \textbf{48.7} & \textbf{87.5} & \textbf{171.6} \\
                & \textcolor{blue}{(+1.9)} & \textcolor{blue}{(+0.3)} & \textcolor{blue}{(+2.2)} & \textcolor{blue}{(+5.8)} \\
                \bottomrule[1.5pt]
            \end{tabular}
        }
        \caption{MSVD dataset}
        \label{tab:main-table1-msvd}
    \end{subtable}

    \begin{subtable}{\columnwidth}
        \centering
        \scalebox{0.83}{
            \begin{tabular}{l cccc}
                \toprule[1.5pt]
                Models & B4 & M & R & C \\  
                \midrule
                *EMCL-Net ~\citep{jin2022expectation} & 45.3 & 30.2 & 63.2 & 54.6 \\
                *CLIP-DCD ~\citep{yang2022clip} & 48.2 & 31.3 & 64.8 & 58.7 \\
                *TextKG ~\citep{gu2023text} & 43.7 & 29.6 & 62.4 & 52.4 \\
                *CoCap ~\citep{shen2023accurate} & 44.4 & 30.3 & 63.4 & 57.2 \\
                *VIOLETv2 ~\citep{fu2023empirical} & - & - & - & 58 \\
                \midrule
                SwinBERT~\citep{lin2022swinbert} & 41.9 & \textbf{29.9} & 62.1 & 53.8 \\
                MAMS + SwinBERT & \textbf{43.3} & 29.8 & \textbf{62.9} & \textbf{54.6} \\
                & \textcolor{blue}{(+1.4)} & \textcolor{blue}{(-0.1)} & \textcolor{blue}{(+0.8)} & \textcolor{blue}{(+0.8)} \\
                \midrule
                mPLUG-2~\citep{xu2023mplug} & 57.9 & \textbf{34.9} & 70.1 & 80.3 \\
                MAMS + mPLUG-2 & \textbf{60.0} & 34.7 & \textbf{71.2} & \textbf{82.9} \\
                & \textcolor{blue}{(+2.1)} & \textcolor{blue}{(-0.2)} & \textcolor{blue}{(+1.1)} & \textcolor{blue}{(+2.6)}\\
                \bottomrule[1.5pt]
            \end{tabular}
        }
        \caption{MSRVTT dataset}
        \label{tab:main-table1-msrvtt}
    \end{subtable}
    \caption{Comparisons of video captioning performances with different captioning models (MSVD and MSRVTT datasets).
    Within the proposed MAMS framework, we used SwinBERT and mPLUG-2.
    The blue numbers in the parenthesis indicate the performance comparison of our MAMS framework with the stand-alone counterparts.
    The asterisk (*) denotes the results reported in the respective paper.}
    \label{tab:main-table1}
\end{table}

\begin{table}[t!]  
    \centering
    \scalebox{0.84}{
        \begin{tabular}{c cccc}
            \toprule[1.5pt]
            Models & B4 & M & R & C\\
            \midrule
            SwinBERT & 9.0 & 15.6 & 37.3 & 109.0 \\
            MAMS + SwinBERT & \textbf{12.5} & \textbf{15.9} & \textbf{40.8} & \textbf{116.7} \\
            & \textcolor{blue}{(+3.5)} & \textcolor{blue}{(+0.3)} & \textcolor{blue}{(+3.5)} & \textcolor{blue}{(+7.7)} \\
            \midrule
            UniVL~\citep{luo2020univl} & 11.2 & 17.6 & 40.1 & 127.0 \\
            MAMS + UniVL & \textbf{14.4} & \textbf{17.8} & \textbf{44.3} & \textbf{133.2} \\
            & \textcolor{blue}{(+3.2)} & \textcolor{blue}{(+0.2)} & \textcolor{blue}{(+4.2)} & \textcolor{blue}{(+6.2)} \\
            \bottomrule[1.5pt]
        \end{tabular}
    }
    \caption{Comparisons of captioning performances with different MAMS models and their stand-alone counterparts (YOUCOOKII dataset).
    The blue numbers in the parenthesis indicate the performance comparison of MAMS models with the stand-alone counterparts.}
    \label{tab:main-table2}
\end{table}

This section discusses the captioning results of a model that integrates the proposed MAMS framework with recent video captioning models.
Tables~\ref{tab:main-table1-msvd}--\ref{tab:main-table1-msrvtt} show that the MAMS significantly improves the video captioning performances of the existing models, SwinBERT and mPLUG-2 across the MSVD and MSRVTT datasets.
The video captioning performance has improved in overall evaluation metrics, and notably, the MAMS framework significantly improves the CIDEr score consistently across different datasets.
Considering that CIDEr is metrics that primarily evaluate how well words are generated, we conjecture that MAMS framework is good at generating words with appropriate meanings.
However, generating precise words could lead to alterations in the structure and order of sentences.
This explains the modest gains in ROUGE and METEOR, metrics sensitive to sentence structure and order. 
Additionally, incorporating the state-of-the-art model mPLUG-2 into MAMS leads to significant improvement.

Table~\ref{tab:main-table2} compares the video captioning performances with different models using the YouCookII dataset.
In particular, we compared the MAMS framework with the corresponding stand-alone counterparts, SwinBERT and UniVL.
Similar to the above claim with the MSVD and MSRVIT datasets, 
the results of Table~\ref{tab:main-table2} demonstrate the outperforming performances of MAMS using SwinBERT and UniVL over stand-alone SwinBERT and UniVL models.
We additionally observe that MAMS can improve the \emph{non}-end-to-end UniVL model.
We conjecture that MAMS framework can be successfully applied to a range of stand-alone video captioning models, improving their performances by a large margin.

\subsection{Ablation Study for the Proposed Framework}

\begin{table}[t!]
    \centering
    \centering
    \scalebox{0.75}{
        \begin{tabular}{cc cccc}
            \toprule[1.5pt]
            Models & Adaptive attn.~mask & B4 & M & R & C \\  
            \midrule
             SwinBERT & $\times$ & 58.2 & 41.3 & 77.5 & 120.6 \\ 
             MAMS + SwinBERT & $\times$ & \textbf{61.3} & 41.6 & 78.6 & 123.5 \\   
             SwinBERT & $\checkmark$ & 61.1 & 41.8 & 78.5 & 122.8  \\
             MAMS + SwinBERT & $\checkmark$ & 60.9 & \textbf{42.1} & \textbf{78.9} & \textbf{125.0} \\ 
            \bottomrule[1.5pt]
        \end{tabular}
    }
    \caption{Performance comparisons between four different configurations of the proposed MAMS framework with SwinBERT (MSVD dataset): 
    the stand-alone model, 
    MAMS without adaptive attention mask,
    the stand-alone model with adaptive attention mask,
    and MAMS with adaptive attention mask. `Adaptive attn. mask' denotes adaptive attention mask.}
    \label{tab:ablation}
\end{table}

This section discusses the ablation study for the proposed MAMS framework.
The second row in Table~\ref{tab:ablation} demonstrates that MAMS can significantly improve the stand-alone counterpart, even without the adaptive attention masking scheme.
It suggests that by adaptively varying the number of frames or visual tokens used for each video, the caption generation performance improves.
The third row in Table~\ref{tab:ablation} demonstrates the effectiveness of the proposed adaptive attention masking scheme.
The adaptive attention masking scheme that is designed for each module of MAMS, 
can be applied to the stand-alone counterpart and significantly improve its captioning performance.
By using the adaptive attention mask, we focus more on visual tokens with a higher contribution in caption generation, resulting in performance improvements.
See the details of implementing the independent integration of an adaptive attention mask into existing models in the supplementary material.
Additionally, the last row in Table~\ref{tab:ablation} implies that MAMS significantly improves the captioning performance by focusing more on essential visual tokens at both the token and frame levels.

\subsection{Analysis of Generated Sentences by the Proposed Framework}

\begin{table}[t!]
    \centering
    \scalebox{0.85}{
        \begin{tabular}{c ccc}
            \toprule[1.5pt]
            MAMS & Adaptive attention mask & Main words & Sub.~words \\
            \midrule
            $\times$ & $\times$ & 28.3 & 48.3 \\ 
            $\times$ & $\checkmark$ & 29.0 & 47.4 \\
            $\checkmark$ & $\times$ & 28.9 & 47.3 \\   
            $\checkmark$ & $\checkmark$ & 29.5 & 48.1 \\ 
            \bottomrule[1.5pt]
        \end{tabular}
    }
    \caption{Words generation performance comparisons using SwinBERT with our MAMS framework with the MSVD dataset. 
    We used BLEU-1 \citep{papineni2002bleu} to measure the performance of both main words and subordinate words results.
    We refer the key components of a sentence, such as the subject, object, complement, and predicate, as `main words.' 
    We refer the remaining words as subordinate words, dubbed `sub words.'
    }
    \label{tab:caption_quality}
\end{table}

This section analyzes the generated sentences by the proposed framework.
Table~\ref{tab:caption_quality} compares the caption generation performances with different combinations of MAMS, adaptive attention mask, and SwinBERT.
Comparing the first and fourth rows in Table~\ref{tab:caption_quality}, 
we explain why does the proposed MAMS framework combined with the proposed adaptive attention masking scheme improve the captioning quality.
The comparisons suggest that the proposed framework allows an existing model to better focus on important visual tokens in generating core words more accurately.

\subsection{Sanity Tests of Module and Token Selectors in MAMS Framework}

\begin{table}[t!]
    \centering
    \begin{subtable}{\columnwidth}
        \centering
        \scalebox{0.85}{
            \begin{tabular}{c cccc}
                \toprule[1.5pt]
                Conditions & B4 & M & R & C \\  
                \midrule
                Eq.~(\ref{eq:mod-select}) & 60.9 & 42.1 & 78.9 & 125.0  \\ 
                Swapped conditions in Eq.~(\ref{eq:mod-select}) & 58.1 & 40.2 & 75.5 & 119.6 \\
                & \textcolor{blue}{(-2.8)} & \textcolor{blue}{(-1.9)} & \textcolor{blue}{(-3.4)} & \textcolor{blue}{(-5.4)}\\
                \bottomrule[1.5pt]
            \end{tabular}
        }
        \caption{Sanity test of module selector}\label{tab:performance_module_selector}
    \end{subtable}

    \begin{subtable}{\columnwidth}
        \centering
        \scalebox{0.85}{
            \begin{tabular}{cc cccc}
                \toprule[1.5pt]
                Token score defs. & B4 & M & R & C \\  
                \midrule
                Eq.~(\ref{eq:token-sig}) & 60.9 & 42.1 & 78.9 & 125.0 \\ 
                $1 - \text{Eq.~(\ref{eq:token-sig})}$  & 55.8 & 37.5 & 75.1 & 116.8 \\
                & \textcolor{blue}{(-5.1)} & \textcolor{blue}{(-4.6)} & \textcolor{blue}{(-3.8)} & \textcolor{blue}{(-8.2)}\\
                \bottomrule[1.5pt]
            \end{tabular}
        }
        \caption{Sanity test of token selector}\label{tab:performance_token_selector}
    \end{subtable}
    \caption{Sanity tests of module and token selectors in the proposed MAMS framework with SwinBERT with the MSVD dataset. 
    See the results with the MSRVTT dataset in the supplementary material.
    The blue numbers in the parentheses indicate the degree of performance degradation.
    }
    \label{tab:performance_module_and_selector}
\end{table}

Table~\ref{tab:performance_module_selector} demonstrates that the proposed module selector in Eq.~(\ref{eq:mod-select}) works appropriately to improve the caption generation performances.
Comparing the performances between MAMS with the inappropriate module selection design (see the second row in Table~\ref{tab:performance_module_selector}) and the stand-alone counterpart (see the first row in Table~\ref{tab:ablation}) show that 
MAMS with the inappropriate module selection design significantly degrades the performances of the stand-alone counterpart.

Table~\ref{tab:performance_token_selector} shows that the proposed token selector using the score defined in Eq.~(\ref{eq:token-sig}) works appropriately to improve the caption generation performances.
Comparing the performances between MAMS with the inappropriate token selector design (see the second row in Table~\ref{tab:performance_token_selector}) and the stand-alone counterpart (see the first row in Table~\ref{tab:ablation}) show that 
MAMS with the inappropriate token selector design significantly degrades the stand-alone counterpart.
The results can justify the token significance score in Eq.~(\ref{eq:token-sig}) in selecting essential tokens in the MAMS framework for accurate caption generations.

\subsection{Analyses for the Proposed Adaptive Attention Masks}

\begin{figure}[t]
    \label{fig:caption results}
  \centering
  \begin{subfigure}{0.43\textwidth}
    \centering
    \includegraphics[width=\linewidth]{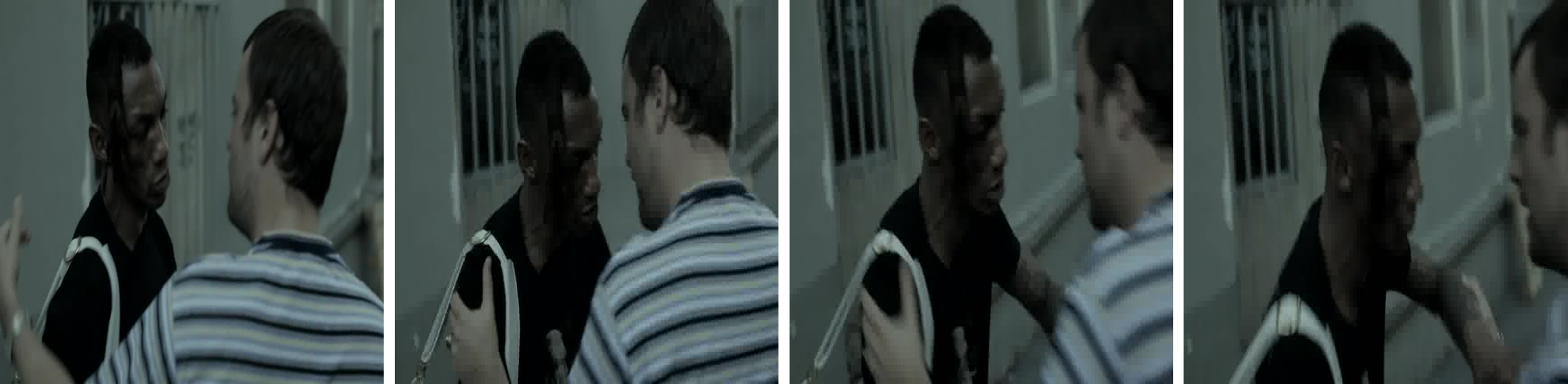}
    \caption{Sampling video frames}
    \label{fig:caption_example_1}
  \end{subfigure}
  
  \begin{subfigure}{0.43\textwidth}
    \centering
    \includegraphics[width=\linewidth]{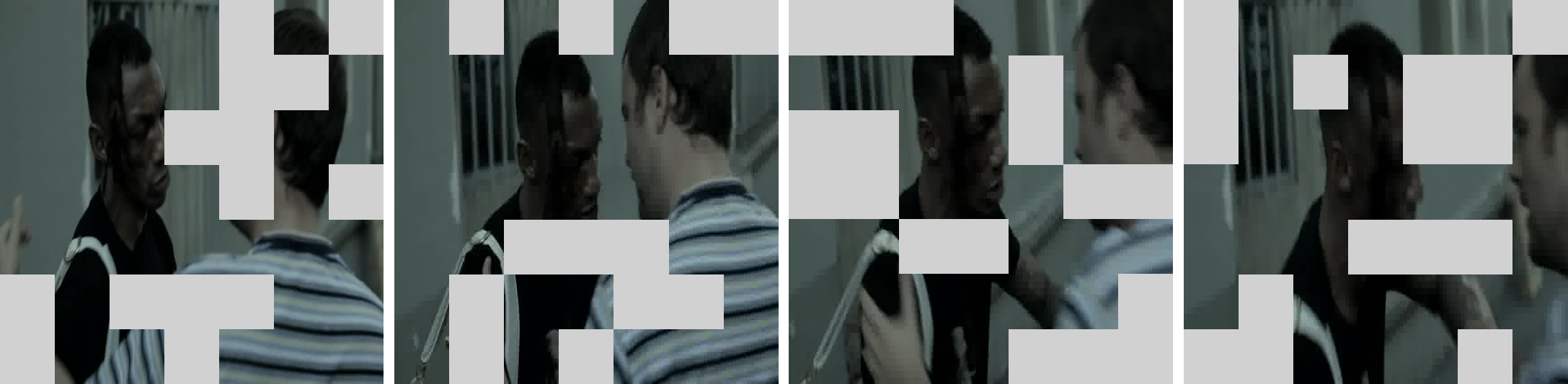}
    \caption{Visualization of unselected visual tokens}
  \label{fig:caption_example_2}
  \end{subfigure}
  
  \begin{subfigure}{0.43\textwidth}
    \centering
    \includegraphics[width=\linewidth]{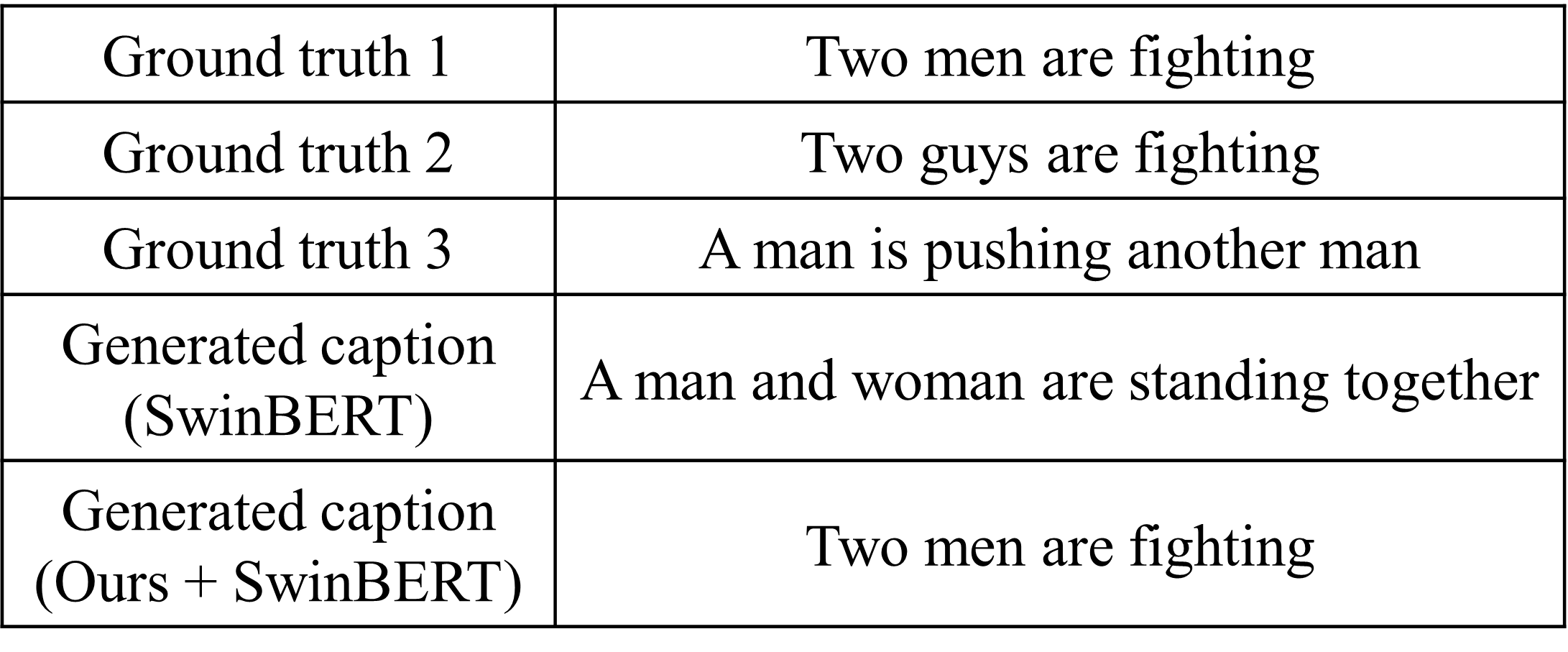}
   \caption{Ground truth captions and generated captions}
  \label{fig:caption_example_3}
  \end{subfigure}
  \caption{Captioning and visualization example with the proposed MAMS framework. (a) shows video frames sampled from \texttt{i3fd4nE8OCI\_174\_181} in the MSVD dataset, (b) provides a visualization of important visual tokens extracted by the adaptive attention mask for each frame, where gray areas represent tokens that are not considered important. (c) presents some ground truth captions from \texttt{i3fd4nE8OCI\_174\_181}, the caption results from SwinBERT, and the caption results of MAMS with SwinBERT.}
  \label{fig:caption_example}
\end{figure}

The proposed adaptive attention masking scheme focuses more on the visual tokens with higher contributions to caption generation, which can be located at the edges, center, or anywhere within the frame, varying for each video.
This section analyzes its effectiveness compared to the fixed learnable mask \citep{lin2022swinbert}, from both the qualitative and quantitative perspectives.

Comparing results in Figure~\ref{fig:caption_example} show that the proposed adaptive attention masking scheme is more reasonable than the fixed learnable attention mask.
In Figure{~\ref{fig:caption_example_3}}, SwinBERT, with the fixed learnable attention mask, fails to understand the man in white clothing at the edges of each frame in Figure~\ref{fig:caption_example_1} and generates an incorrect word, `woman'.
In Figure{~\ref{fig:caption_example_2}}, the adaptive attention masks correctly select the parts with the man in white clothing and the man in black clothing as important visual tokens for caption generation, resulting in accurate captions as shown in Figure{~\ref{fig:caption_example_3}}.

Table~\ref{tab:ablation_mask} shows that the adaptive attention masking scheme outperforms the fixed learnable attention mask in captioning performance, both with the MAMS framework+SwinBERT and the stand-alone counterpart.
These results imply that the proposed adaptive attention masks, while overcoming the limitations of the fixed learnable attention mask, better focus on essential tokens for caption generation.

\begin{table}[t!]
    \centering
    \centering
    \scalebox{0.77}{
        \begin{tabular}{cc cccc}
            \toprule[1.5pt]
            Models & Mask & B4 & M & R & C \\  
            \midrule
             SwinBERT & Fixed attn.~mask & 58.2 & 41.3 & 77.5 & 120.6 \\ 
             SwinBERT & Adaptive attn.~mask & \textbf{61.1} & 41.8 & 78.5 & 122.8 \\   
             MAMS + SwinBERT & Fixed attn.~mask & 60.2 & 41.4 & 78.2 & 123.3  \\
             MAMS + SwinBERT & Adaptive attn.~mask & 60.9 & \textbf{42.1} & \textbf{78.9} & \textbf{125.0} \\ 
            \bottomrule[1.5pt]
        \end{tabular}
    }
    \caption{Captioning performance comparisons with the MSVD dataset between SwinBERT and the MAMS framework with SwinBERT, with either a fixed learnable attention mask or our adaptive attention mask applied to each model, where Fixed attn. mask and Adaptive attn. mask refers to a fixed learnable attention mask and an adaptive attention mask, respectively.
    }
    \label{tab:ablation_mask}
\end{table}

\subsection{Comparisons with Different Numbers of Generation Modules}\label{sec:different flow counts}

\begin{table}
    \centering
    \centering
    \scalebox{0.83}{
        \begin{tabular}{c cccc}
            \toprule[1.5pt]
            Models & B4 & M & R & C \\  
            \midrule
            SwinBERT & 58.2 & 41.3 & 77.5 & 120.6 \\ 
            Ours + SwinBERT -- default & \textbf{60.9} & \textbf{42.1} & \textbf{78.9} & \textbf{125.0} \\
            Ours + SwinBERT -- 3 candidates & 59.1 & 40.8 & 77.5 & 119.5 \\
            Ours + SwinBERT -- 4 candidates & 57.4 & 39.2 &75.1 & 118.1 \\
            \bottomrule[1.5pt]
        \end{tabular}
    }
    \caption{Performance comparisons of our MAMS framework) with different numbers of candidate captioning modules for MSVD dataset. The default setup in MAMS uses two generation module candidates; see Figure~\ref{fig:overall}.
    See results with the MSRVTT dataset in the supplementary material.
    }
    \label{tab:multi-flow}
\end{table}

The default configuration in our MAMS framework selects between two caption generation modules: a large module and a small module.
This section compares the performance of MAMS with different numbers of generation modules.
Table~\ref{tab:multi-flow} compares captioning performances of MAMS incorporating SwinBERT by varying the number of generation module candidates.
It shows that the default setup selecting between two generation modules outperforms configurations that select among three or four generation modules, each handling different numbers of frames.
Increasing the number of candidate modules can divide the training data into a larger number of groups, potentially resulting in some generation modules having a limited number of training samples, which leads to performance degradation.
Unless the training data contains a large number of samples sufficient to adequately train all modules, the default setup is likely to be preferred.

\section{Conclusion}
In this paper, we propose the first model-agnostic framework in video captioning, that selects a caption generation module of appropriate size for each video.
To further enhance the video captioning performance, we propose a new adaptive attention masking scheme for the MAMS framework by focusing on more significant visual tokens, which can guide in identifying the main words.
Our numerical experiments across different datasets demonstrate that the proposed MAMS framework significantly and consistently improves the recent video captioning models.

For future work, we plan to further improve captioning performances and gain further insights by focusing more on important textual tokens, as well as important visual tokens.
Additionally, we aim to extend the underlying principles of the MAMS framework to other video understanding tasks, such as video summarization and video question answering, to broaden its applicability and impact in the field of video analysis.

\section{Acknowledgements}

This work was supported by the Institute of Information \& Communications Technology Planning \& evaluation (IITP) and the National Research Foundation of Korea (NRF) grants funded by the Korea government (MSIT) (RS-2019-II190421 (1\%), IITP-2025-RS-2020-II201821 (1\%), NRF-2021M3H4A1A02056037 (8\%), RS-2024-00438686 (10\%), RS-2024-00436936 (10\%), RS-2024-00360227 (10\%), RS-2024-00448809 (10\%), 2022R1F1A1074546 (10\%), and RS-2023-00213455 (10\%), IBS-R015-D1 (10\%), and Korea Institute for Advancement of Technology (KIAT) grants funded by the Korea government (MOTIE) (RS-2024-00418086 (10\%) and P0022098 (10\%)).  


\bibliography{aaai25}

\begin{thebibliography}{37}
\providecommand{\natexlab}[1]{#1}

\bibitem[{Aafaq et~al.(2019)Aafaq, Akhtar, Liu, Gilani, and
  Mian}]{aafaq2019spatio}
Aafaq, N.; Akhtar, N.; Liu, W.; Gilani, S.~Z.; and Mian, A. 2019.
\newblock Spatio-temporal dynamics and semantic attribute enriched visual
  encoding for video captioning.
\newblock In \emph{Proceedings of the IEEE/CVF Conference on Computer Vision
  and Pattern Recognition}, 12487--12496.

\bibitem[{Arnab et~al.(2021)Arnab, Dehghani, Heigold, Sun, Lu{\v{c}}i{\'c}, and
  Schmid}]{arnab2021vivit}
Arnab, A.; Dehghani, M.; Heigold, G.; Sun, C.; Lu{\v{c}}i{\'c}, M.; and Schmid,
  C. 2021.
\newblock Vivit: A video vision transformer.
\newblock In \emph{Proceedings of the IEEE/CVF International Conference on
  Computer Vision}, 6836--6846.

\bibitem[{Bahdanau, Cho, and Bengio(2015)}]{bahdanau2014neural}
Bahdanau, D.; Cho, K.; and Bengio, Y. 2015.
\newblock Neural machine translation by jointly learning to align and
  translate.
\newblock In \emph{Proceedings of the International Conference on Learning
  Representations}.

\bibitem[{Banerjee et~al.(2005)Banerjee, Satanjeev, Lavie, and
  Alon}]{banerjee2005meteor}
Banerjee; Satanjeev; Lavie; and Alon. 2005.
\newblock METEOR: An automatic metric for MT evaluation with improved
  correlation with human judgments.
\newblock In \emph{Proceedings of the Association for Computational
  Linguistics}, 65--72.

\bibitem[{Cao et~al.(2023)Cao, Huang, Liao, and Mao}]{cao2023ada}
Cao, Q.; Huang, H.; Liao, M.; and Mao, X. 2023.
\newblock Ada-SwinBERT: Adaptive Token Selection for Efficient Video Captioning
  with Online Self-Distillation.
\newblock In \emph{Proceedings of the IEEE International Conference on
  Multimedia and Expo}, 7--12.

\bibitem[{Chen et~al.(2011)Chen, David, Dolan, and B}]{msvd}
Chen; David; Dolan; and B, W. 2011.
\newblock Collecting highly parallel data for paraphrase evaluation.
\newblock In \emph{Proceedings of the Association for Computational
  Linguistics}, 190--200.

\bibitem[{Chen et~al.(2023)Chen, Li, Wang, Zhao, Sun, Zhu, and
  Liu}]{chen2023vast}
Chen, S.; Li, H.; Wang, Q.; Zhao, Z.; Sun, M.; Zhu, X.; and Liu, J. 2023.
\newblock VAST: A Vision-Audio-Subtitle-Text Omni-Modality Foundation Model and
  Dataset.
\newblock In \emph{Advances in Neural Information Processing Systems}.

\bibitem[{Das et~al.(2013)Das, Xu, Doell, and Corso}]{das2013thousand}
Das, P.; Xu, C.; Doell, R.~F.; and Corso, J.~J. 2013.
\newblock A thousand frames in just a few words: Lingual description of videos
  through latent topics and sparse object stitching.
\newblock In \emph{Proceedings of the IEEE Conference on Computer Vision and
  Pattern Recognition}, 2634--2641.

\bibitem[{Fu et~al.(2023)Fu, Li, Gan, Lin, Wang, Wang, and
  Liu}]{fu2023empirical}
Fu, T.-J.; Li, L.; Gan, Z.; Lin, K.; Wang, W.~Y.; Wang, L.; and Liu, Z. 2023.
\newblock An empirical study of end-to-end video-language transformers with
  masked visual modeling.
\newblock In \emph{Proceedings of the IEEE/CVF Conference on Computer Vision
  and Pattern Recognition}, 22898--22909.

\bibitem[{Gao et~al.(2023)Gao, Yizhao, Lu, and Zhiwu}]{Gao_2022_ACCV}
Gao; Yizhao; Lu; and Zhiwu. 2023.
\newblock SST-VLM: Sparse Sampling-Twice Inspired Video-Language Model.
\newblock In \emph{Proceedings of the Asian Conference on Computer Vision},
  537–553.

\bibitem[{Gu et~al.(2023)Gu, Chen, Wang, Zhang, Luo, and Wen}]{gu2023text}
Gu, X.; Chen, G.; Wang, Y.; Zhang, L.; Luo, T.; and Wen, L. 2023.
\newblock Text with Knowledge Graph Augmented Transformer for Video Captioning.
\newblock In \emph{Proceedings of the IEEE/CVF Conference on Computer Vision
  and Pattern Recognition}, 18941--18951.

\bibitem[{Jang et~al.(2017)Jang, Eric, Gu, Shixiang, Poole, and
  Ben}]{jang2016categorical}
Jang; Eric; Gu; Shixiang; Poole; and Ben. 2017.
\newblock Categorical reparameterization with gumbel-softmax.
\newblock In \emph{Proceedings of the International Conference on Learning
  Representations}.

\bibitem[{Jin et~al.(2022)Jin, Huang, Liu, Wu, Ge, Song, Clifton, and
  Chen}]{jin2022expectation}
Jin, P.; Huang, J.; Liu, F.; Wu, X.; Ge, S.; Song, G.; Clifton, D.; and Chen,
  J. 2022.
\newblock Expectation-maximization contrastive learning for compact
  video-and-language representations.
\newblock \emph{Advances in Neural Information Processing Systems}, 35:
  30291--30306.

\bibitem[{Kojima et~al.(2002)Kojima, Atsuhiro, Tamura, Takeshi, Fukunaga, and
  Kunio}]{kojima2002natural}
Kojima; Atsuhiro; Tamura; Takeshi; Fukunaga; and Kunio. 2002.
\newblock Natural language description of human activities from video images
  based on concept hierarchy of actions.
\newblock \emph{International Journal of Computer Vision}, 50: 171--184.

\bibitem[{Kuo et~al.(2023)Kuo, Piergiovanni, Kim, xiyang luo, Caine, Li, Ogale,
  Zhou, Dai, Chen, Cui, and Angelova}]{kuo2023mammut}
Kuo, W.; Piergiovanni, A.; Kim, D.; xiyang luo; Caine, B.; Li, W.; Ogale, A.;
  Zhou, L.; Dai, A.~M.; Chen, Z.; Cui, C.; and Angelova, A. 2023.
\newblock Ma{MMUT}: A Simple Architecture for Joint Learning for MultiModal
  Tasks.
\newblock \emph{Transactions on Machine Learning Research}.

\bibitem[{Li et~al.(2021)Li, Lei, Gan, Yu, Chen, Pillai, Zhou, Wang, Wang
  et~al.}]{li2021value}
Li, L.; Lei, J.; Gan, Z.; Yu, L.; Chen, Y.-C.; Pillai, Y., Rohitand~Cheng;
  Zhou, L.; Wang, X.~E.; Wang, W.~Y.; et~al. 2021.
\newblock VALUE: A Multi-Task Benchmark for Video-and-Language Understanding
  Evaluation.
\newblock \emph{Advances in Neural Information Processing Systems}.

\bibitem[{Liang et~al.(2022)Liang, Ge, Tong, Song, Wang, and
  Xie}]{liang2022evit}
Liang, Y.; Ge, C.; Tong, Z.; Song, Y.; Wang, J.; and Xie, P. 2022.
\newblock Not All Patches are What You Need: Expediting Vision Transformers via
  Token Reorganizations.
\newblock In \emph{Proceedings of the International Conference on Learning
  Representations}.

\bibitem[{Lin et~al.(2004)Lin, Chin-Yew, Och, and Josef}]{rouge}
Lin; Chin-Yew; Och; and Josef, F. 2004.
\newblock Automatic evaluation of machine translation quality using longest
  common subsequence and skip-bigram statistics.
\newblock In \emph{Proceedings of Association for Computational Linguistics},
  605--612.

\bibitem[{Lin et~al.(2022)Lin, Li, Lin, Ahmed, Gan, Liu, Lu, and
  Wang}]{lin2022swinbert}
Lin, K.; Li, L.; Lin, C.-C.; Ahmed, F.; Gan, Z.; Liu, Z.; Lu, Y.; and Wang, L.
  2022.
\newblock Swinbert: End-to-end transformers with sparse attention for video
  captioning.
\newblock In \emph{Proceedings of the IEEE/CVF Conference on Computer Vision
  and Pattern Recognition}, 17949--17958.

\bibitem[{Liu et~al.(2023)Liu, Xiangcheng, Wu, Tianyi, Guo, and
  Guodong}]{Ada-Vit}
Liu; Xiangcheng; Wu; Tianyi; Guo; and Guodong. 2023.
\newblock Adaptive sparse ViT: towards learnable adaptive token pruning by
  fully exploiting self-attention.
\newblock In \emph{Proceedings of the Thirty-Second International Joint
  Conference on Artificial Intelligence}.

\bibitem[{Luo et~al.(2020)Luo, Ji, Shi, Huang, Duan, Li, Li, Bharti, and
  Zhou}]{luo2020univl}
Luo, H.; Ji, L.; Shi, B.; Huang, H.; Duan, N.; Li, T.; Li, J.; Bharti, T.; and
  Zhou, M. 2020.
\newblock Univl: A unified video and language pre-training model for multimodal
  understanding and generation.
\newblock \emph{arXiv preprint arXiv:2002.06353}.

\bibitem[{Pan et~al.(2020)Pan, Cai, Huang, Lee, Gaidon, Adeli, and
  Niebles}]{pan2020spatio}
Pan, B.; Cai, H.; Huang, D.-A.; Lee, K.-H.; Gaidon, A.; Adeli, E.; and Niebles,
  J.~C. 2020.
\newblock Spatio-temporal graph for video captioning with knowledge
  distillation.
\newblock In \emph{Proceedings of the IEEE/CVF Conference on Computer Vision
  and Pattern Recognition}, 10870--10879.

\bibitem[{Papineni et~al.(2002)Papineni, Roukos, Ward, and
  Zhu}]{papineni2002bleu}
Papineni, K.; Roukos, S.; Ward, T.; and Zhu, W.-J. 2002.
\newblock Bleu: a method for automatic evaluation of machine translation.
\newblock In \emph{Proceedings of the Association for Computational
  Linguistics}, 311--318.

\bibitem[{Paszke et~al.(2019)Paszke, Gross, Massa, Lerer, Bradbury, Chanan,
  Killeen, Lin, Gimelshein, Antiga et~al.}]{paszke2019pytorch}
Paszke, A.; Gross, S.; Massa, F.; Lerer, A.; Bradbury, J.; Chanan, G.; Killeen,
  T.; Lin, Z.; Gimelshein, N.; Antiga, L.; et~al. 2019.
\newblock Pytorch: An imperative style, high-performance deep learning library.
\newblock \emph{Advances in Neural Information Processing Systems}, 32.

\bibitem[{Pei et~al.(2019)Pei, Zhang, Wang, Ke, Shen, and Tai}]{pei2019memory}
Pei, W.; Zhang, J.; Wang, X.; Ke, L.; Shen, X.; and Tai, Y.-W. 2019.
\newblock Memory-attended recurrent network for video captioning.
\newblock In \emph{Proceedings of the IEEE/CVF Conference on Computer Vision
  and Pattern Recognition}, 8347--8356.

\bibitem[{Shen et~al.(2023)Shen, Gu, Xu, Fan, Wen, and
  Zhang}]{shen2023accurate}
Shen, Y.; Gu, X.; Xu, K.; Fan, H.; Wen, L.; and Zhang, L. 2023.
\newblock Accurate and Fast Compressed Video Captioning.
\newblock In \emph{Proceedings of the IEEE/CVF International Conference on
  Computer Vision}.

\bibitem[{Shi et~al.(2020)Shi, Ji, Niu, Duan, Zhou, and Chen}]{shi2020learning}
Shi, B.; Ji, L.; Niu, Z.; Duan, N.; Zhou, M.; and Chen, X. 2020.
\newblock Learning semantic concepts and temporal alignment for narrated video
  procedural captioning.
\newblock In \emph{Proceedings of the ACM international conference on
  multimedia}, 4355--4363.

\bibitem[{Sun et~al.(2019)Sun, Myers, Vondrick, Murphy, and
  Schmid}]{sun2019videobert}
Sun, C.; Myers, A.; Vondrick, C.; Murphy, K.; and Schmid, C. 2019.
\newblock Videobert: A joint model for video and language representation
  learning.
\newblock In \emph{Proceedings of the IEEE/CVF International Conference on
  Computer Vision}, 7464--7473.

\bibitem[{Sutskever et~al.(2014)Sutskever, Ilya, Vinyals, Oriol, Le, and
  V}]{sutskever2014sequence}
Sutskever; Ilya; Vinyals; Oriol; Le; and V, Q. 2014.
\newblock Sequence to sequence learning with neural networks.
\newblock \emph{Advances in Neural Information Processing Systems}, 27.

\bibitem[{Vedantam et~al.(2015)Vedantam, Ramakrishna, Lawrence~Zitnick, Parikh,
  and Devi}]{vedantam2015cider}
Vedantam; Ramakrishna; Lawrence~Zitnick, C.; Parikh; and Devi. 2015.
\newblock Cider: Consensus-based image description evaluation.
\newblock In \emph{Proceedings of the IEEE Conference on Computer Vision and
  Pattern Recognition}, 4566--4575.

\bibitem[{Wang et~al.(2019)Wang, Ma, Zhang, Jiang, Wang, and
  Liu}]{wang2019controllable}
Wang, B.; Ma, L.; Zhang, W.; Jiang, W.; Wang, J.; and Liu, W. 2019.
\newblock Controllable video captioning with pos sequence guidance based on
  gated fusion network.
\newblock In \emph{Proceedings of the IEEE/CVF International Conference on
  Computer Vision}, 2641--2650.

\bibitem[{Wang et~al.(2022)Wang, Li, Xu, Zhou, Lei, Lin, Wang, Yang, Zhu, Hoiem
  et~al.}]{wang2022language}
Wang, Z.; Li, M.; Xu, R.; Zhou, L.; Lei, J.; Lin, X.; Wang, S.; Yang, Z.; Zhu,
  C.; Hoiem, D.; et~al. 2022.
\newblock Language models with image descriptors are strong few-shot
  video-language learners.
\newblock \emph{Advances in Neural Information Processing Systems}, 35:
  8483--8497.

\bibitem[{Xu et~al.(2023)Xu, Ye, Yan, Shi, Ye, Xu, Li, Bi, Qian, Wang
  et~al.}]{xu2023mplug}
Xu, H.; Ye, Q.; Yan, M.; Shi, Y.; Ye, J.; Xu, Y.; Li, C.; Bi, B.; Qian, Q.;
  Wang, W.; et~al. 2023.
\newblock mplug-2: A modularized multi-modal foundation model across text,
  image and video.
\newblock In \emph{Proceedings of the International Conference on Machine
  Learning}.

\bibitem[{Xu et~al.(2016)Xu, Mei, Yao, and Rui}]{msrvtt}
Xu, J.; Mei, T.; Yao, T.; and Rui, Y. 2016.
\newblock Msr-vtt: A large video description dataset for bridging video and
  language.
\newblock In \emph{Proceedings of the IEEE Conference on Computer Vision and
  Pattern Recognition}, 5288--5296.

\bibitem[{Yang et~al.(2022)Yang, Bang, Zhan, Tong, Zou, and
  Yuexian}]{yang2022clip}
Yang; Bang; Zhan; Tong; Zou; and Yuexian. 2022.
\newblock CLIP meets video captioning: Concept-aware representation learning
  does matter.
\newblock In \emph{Proceedings of the Chinese Conference on Pattern Recognition
  and Computer Vision}, 368--381.

\bibitem[{Yang et~al.(2023)Yang, Nagrani, Seo, Miech, Pont-Tuset, Laptev,
  Sivic, and Schmid}]{yang2023vid2seq}
Yang, A.; Nagrani, A.; Seo, P.~H.; Miech, A.; Pont-Tuset, J.; Laptev, I.;
  Sivic, J.; and Schmid, C. 2023.
\newblock Vid2seq: Large-scale pretraining of a visual language model for dense
  video captioning.
\newblock In \emph{Proceedings of the IEEE/CVF Conference on Computer Vision
  and Pattern Recognition}, 10714--10726.

\bibitem[{Zhou et~al.(2018)Zhou, Luowei, Xu, Chenliang, Corso, and
  Jason}]{youcook}
Zhou; Luowei; Xu; Chenliang; Corso; and Jason. 2018.
\newblock Towards automatic learning of procedures from web instructional
  videos.
\newblock In \emph{Proceedings of the AAAI Conference on Artificial
  Intelligence}, volume~32.

\end{thebibliography}

\end{document}